\pdfoutput=1

\documentclass[sigconf,natbib,9pt]{acmart}

\copyrightyear{2017}
\acmYear{2017}
\setcopyright{rightsretained}
\acmConference{SIGIR 2017 Workshop on Neural Information Retrieval (Neu-IR'17)}{}{August 7--11, 2017, Shinjuku, Tokyo, Japan}
\acmPrice{}
\usepackage{amsmath}
\usepackage{booktabs}
\usepackage{color}
\usepackage{enumitem}
\usepackage{footnote}
\usepackage{hyperref}
\usepackage{grffile}
\usepackage{mathtools}

\usepackage{multirow}
\usepackage[np, autolanguage]{numprint}
\usepackage[nodisplayskipstretch]{setspace}
\usepackage{subfig}
\usepackage{tabularx}
\usepackage{tikz}
\usetikzlibrary{arrows, arrows.meta, calc, decorations.markings, decorations.pathreplacing, decorations.pathmorphing, fadings, patterns, positioning, positioning, shapes, shapes.misc}
\usepackage{todonotes}
\usepackage{paralist}
\usepackage[labelfont=bf,textfont=bf]{caption}
\usepackage{etoolbox}
\usepackage{array}
\usepackage{xintfrac}
\usepackage{acronym}
\usepackage[normalem]{ulem}
\usepackage[frozencache]{minted}

\newfloat{algorithm}{t}{lop}
\floatname{algorithm}{Code snippet}

\usepackage{etoolbox}
\AtBeginEnvironment{minted}{\singlespacing%
    \fontsize{8}{8}\selectfont}
\newminted{python}{mathescape, numbersep=5pt, autogobble, framesep=2mm, fontsize=\footnotesize}

\setcitestyle{square,comma,numbers,sort&compress,sectionbib}

\hypersetup{pdfinfo={
Title={Semantic Entity Retrieval Toolkit},
Author={Christophe Van Gysel, Maarten de Rijke, Evangelos Kanoulas}
}}

\def \paperImplementationUrl {https://github.com/cvangysel/SERT}

\allowdisplaybreaks

\begin{document}

\nocite{VanGysel2016experts,VanGysel2016products}

\title{Semantic Entity Retrieval Toolkit}
\titlenote{The toolkit is licensed under the permissive MIT open-source license and can be found at \url{\paperImplementationUrl}.}

\author{Christophe Van Gysel}
\orcid{0000-0003-3433-7317}
\affiliation{%
\institution{University of Amsterdam}
\city{Amsterdam}
\country{The Netherlands}
}
\email{cvangysel@uva.nl}

\author{Maarten de Rijke}
\orcid{0000-0002-1086-0202}
\affiliation{%
\institution{University of Amsterdam}
\city{Amsterdam}
\country{The Netherlands}
}
\email{derijke@uva.nl}

\author{Evangelos Kanoulas}
\orcid{0000-0002-8312-0694}
\affiliation{%
\institution{University of Amsterdam}
\city{Amsterdam}
\country{The Netherlands}
}
\email{e.kanoulas@uva.nl}

\renewcommand{\shortauthors}{Van Gysel et al.}

\acrodef{SERT}{Semantic Entity Retrieval Toolkit}

\newcommand{\SERT}{\acs{SERT}}

\begin{abstract}
Unsupervised learning of low-dimensional, semantic representations of words and entities has recently gained attention. In this paper we describe the Semantic Entity Retrieval Toolkit (\SERT{}) that provides implementations of our previously published entity representation models. The toolkit provides a unified interface to different representation learning algorithms, fine-grained parsing configuration and can be used transparently with GPUs. In addition, users can easily modify existing models or implement their own models in the framework. After model training, \SERT{} can be used to rank entities according to a textual query and extract the learned entity/word representation for use in downstream algorithms, such as clustering or recommendation.
\end{abstract}


\maketitle

\section{Introduction}

The unsupervised learning of low-dimensional, semantic representations of words and entities has recently gained attention for the entity-oriented tasks of expert finding \citep{VanGysel2016experts} and product search \citep{VanGysel2016products}. Representations are learned from a document collection and domain-specific associations between documents and entities. Expert finding is the task of finding the right person with the appropriate skills or knowledge \citep{Balog2012survey} and an association indicates document authorship (e.g., academic papers) or involvement in a project (e.g., annual progress reports). In the case of product search, an associated document is a product description or review \citep{VanGysel2016products}.

In this paper we describe the Semantic Entity Retrieval Toolkit (\SERT{}) that provides implementations of our previously published entity representation models \citep{VanGysel2016experts,VanGysel2016products}. Beyond a unified interface that combines different models, the toolkit allows for fine-grained parsing configuration and GPU-based training through integration with Theano \citep{Lasagne2015,Theano2016}. Users can easily extend existing models or implement their own models within the unified framework. After model training, \SERT{} can compute matching scores between an entity and a piece of text (e.g., a query). This matching score can then be used for ranking entities, or as a feature in a downstream machine learning system, such as the learning to rank component of a search engine. In addition, the learned representations can be extracted and used as feature vectors in entity clustering or recommendation tasks \citep{VanGysel2017expertregularities}. The toolkit is licensed under the permissive MIT open-source license; see the footnote on the first page.
\section{The Toolkit}

\begin{figure}
\centering
\begin{tikzpicture}
\node [draw, text width=2.5cm, align=center] (prepare) at (-0.65, 0) {\textbf{Text processing}\\(prepare; \S\ref{sec:prepare})};
\node [draw, text width=2.2cm, align=center] (train) at (2.25, 0) {\textbf{Repr. learning}\\(train; \S\ref{sec:train})};
\node [draw, text width=1.8cm, align=center] (query) at (5, 0) {\textbf{Inference}\\(query; \S\ref{sec:query})};

\node [above = 0.10cm of train] {\textbf{\acl{SERT}}};

\draw [->, line width=0.25mm] (prepare) -- (train);
\draw [->, line width=0.25mm] (train) -- (query);
\end{tikzpicture}

\caption{Schematic overview of the different pipeline components of SERT. The collection is parsed, processed and packaged in a numerical format using the \emph{prepare} (\S\ref{sec:prepare}) utility. Afterwards, the \emph{training} (\S\ref{sec:train}) utility learns representations of entities and words and the \emph{query} (\S\ref{sec:query}) utility is used to compute matching scores between entities and queries.\label{fig:overview}}
\vspace*{-\baselineskip}
\end{figure}
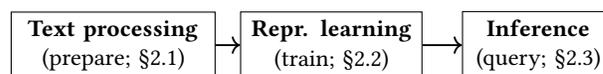

\SERT{} is organized as a pipeline of utilities as depicted in Fig.~\ref{fig:overview}. First, a collection of documents and entity associations is processed and packaged using a numerical format (\S\ref{sec:prepare}). Low-dimensional representations of words and entities are then learned (\S\ref{sec:train}) and afterwards the representations can be used to make inferences (\S\ref{sec:query}).

\subsection{Collection parsing and preparation}
\label{sec:prepare}

To begin, \SERT{} constructs a vocabulary that will be used to tokenize the document collection. Non-significant words that are too frequent (e.g., stopwords), noisy (e.g., single characters) and rare words are filtered out. Words that do not occur in the dictionary are ignored. Afterwards, word sequences are extracted from the documents and stored together with the associated entities in the numerical format provided by NumPy \citep{VanDerWalt2011numpy}. Word sequences can be extracted consecutively or a stride can be specified to extract non-consecutive windows. In addition, a hierarchy of word sequence extractors can be applied to extract skip-grams, i.e., word sequences where a number of tokens are skipped after selecting a token \citep{Guthrie2006skipgram}. To support short documents, a special-purpose padding token can be used to fill up word sequences that are longer than a particular document.

After word sequence extraction, a weight can be assigned to each word sequence/entity pair that can be used to re-weight the training objective. For example, in the case of expert finding \citep{VanGysel2016experts}, this weight is the reciprocal of the document length of the document where the sequence was extracted from. This avoids a bias in the objective towards long documents. An alternative option that exists within the toolkit is to resample word sequence/entity pairs such that every entity is associated with the same number of word sequences, as used for product search \citep{VanGysel2016products}.

\subsection{Representation learning}
\label{sec:train}

\begin{algorithm}
\begin{minted}[frame=lines]{python}
class ExampleModel(VectorSpaceLanguageModelBase):

    def __init__(self, *args, **kwargs):
        super(ExampleModel, self).__init__(
            *args, **kwargs)

        # Define model architecture.
        input_layer = InputLayer(
            shape=(self.batch_size, self.window_size))

        ...

        def loss_fn(pred, actual, _):
            # Compute symbolic loss between
            # predicted/actual entities.

        # The framework deals with underlying boilerplate.
        self._finalize(loss_fn, ....)

    def get_representations(self):
        # Returns the representations and parameters
        # to be extracted.
\end{minted}
\caption{Illustrative example of the \SERT{} model interface. The full interface supports more functionality omitted here for brevity. Users can define a symbolic graph of computation using the Theano library \citep{Theano2016} in combination with Lasagne \citep{Lasagne2015}.\label{snippet:model}}
\vspace*{-\baselineskip}
\end{algorithm}

After the collection has been processed and packaged in a machine-friendly format, representations of words and entities can be learned. The toolkit includes implementations of state-of-the-art representation learning models that were applied to expert finding \citep{VanGysel2016experts} and product search \citep{VanGysel2016products}.
Users of the toolkit can use these implementations to learn representations out-of-the-box or adapt the algorithms to their needs. In addition, users can implement their own models by extending an interface provided by the framework. Code snippet~\ref{snippet:model} shows an example of a model implemented in the \SERT{} toolkit where users can define a symbolic cost function that will be optimized using Theano \citep{Theano2016}. Due to the component-wise organization of the toolkit (Fig.~\ref{fig:overview}), modeling and text processing are separated from each other. Consequently, researchers can focus on modeling and representation learning only. In addition, any improvements to the collection processing (\S\ref{sec:prepare}) collectively benefits all models implemented in \SERT{}.

\subsection{Entity ranking \& other uses of the representations}
\label{sec:query}

Once a model has been trained, \SERT{} can be used to rank entities w.r.t. a textual query. The concrete implementation used to rank entities depends on the model that was trained. In the most generic case, a matching score is computed for every entity and entities are ranked in decreasing order of his score. However, in the special case when the model is interpreted as a metric vector space \citep{VanGysel2016products,Boytsov2016knn}, \SERT{} casts entity ranking as a $k$-nearest neighbor problem and uses specialized data structures for retrieval \citep{Kibriya2007knn}. After ranking, \SERT{} outputs the entity rankings as a TREC-compatible file that can be used as input to the \texttt{trec\_eval}\footnote{\url{https://github.com/usnistgov/trec_eval}} evaluation utility.

Apart from entity ranking, the learned representations and model-specific parameters can be extracted conveniently from the models through the interface\footnote{See \texttt{get\_representations} in Snippet~\ref{snippet:model}.} and used for down-stream tasks such as clustering, recommendation and determining entity importance as shown in \citep{VanGysel2017expertregularities}.

\section{Conclusions}

In this paper we described the \acl{SERT}, a toolkit that learns latent representations of words and entities. The toolkit contains implementations of state-of-the-art entity representations algorithms \citep{VanGysel2016products,VanGysel2016experts} and consists of three components: text processing, representation learning and inference. Users of the toolkit can easily make changes to existing model implementations or contribute their own models by extending an interface provided by the \SERT{} framework.

Future work includes integration with Pyndri \citep{VanGysel2017pyndri} such that document collections indexed with Indri can transparently be used to train entity representations. In addition, integration with machine learning frameworks besides Theano, such as TensorFlow and PyTorch, will make it easier to integrate existing models into \SERT{}.

\smallskip
\begin{spacing}{1}
\noindent\small
\textbf{Acknowledgments.}
The authors would like to thank the anonymous reviewers for their valuable comments and suggestions.
This research was supported by
Ahold Delhaize,
Amsterdam Data Science,
the Bloomberg Research Grant program,
the Criteo Faculty Research Award program,
the Dutch national program COMMIT,
Elsevier,
the European Community's Seventh Framework Programme (FP7/2007-2013) under
grant agreement nr 312827 (VOX-Pol),
the Google Faculty Research Award scheme,
the Microsoft Research Ph.D.\ program,
the Netherlands Institute for Sound and Vision,
the Netherlands Organisation for Scientific Research (NWO)
under pro\-ject nrs
612.001.116, 
HOR-11-10, 
CI-14-25, 
652.\-002.\-001, 
612.\-001.\-551, 
652.\-001.\-003, 
and
Yandex.
All content represents the opinion of the authors, which is not necessarily shared or endorsed by their respective employers and/or sponsors.
\end{spacing}

\bibliographystyle{abbrvnatnourl}
\bibliography{neuir2017-sert}

\end{document}